\documentclass[10pt,twocolumn,letterpaper]{article}

\usepackage{cvpr}
\usepackage{times}
\usepackage{epsfig}
\usepackage{graphicx}
\usepackage{amsmath}
\usepackage{amssymb}

\usepackage[breaklinks=true,bookmarks=false]{hyperref}

\cvprfinalcopy % *** Uncomment this line for the final submission

\def\cvprPaperID{****} % *** Enter the CVPR Paper ID here

\begin{document}
%%%%%%%%% TITLE
\title{Multi-hierarchical Independent Correlation Filters for Visual Tracking}

\author{Shuai Bai\thanks{Both authors contributed equally.}   $^{\,,}$$^1$, Zhiqun He$^{*,1}$, Ting-Bing Xu$^{2}$, Zheng Zhu$^{2}$, Yuan Dong$^{1}$, Hongliang Bai$^{3}$\\
\textsuperscript{1}Beijing University of Posts and Telecommunications\\
\textsuperscript{2}Institute of Automation of Chinese Academy of Sciences \textsuperscript{3}Beijing FaceAll Co.\\
{\tt\small \{baishuai, he010103, yuandong\}@bupt.edu.cn tingbing.xu@nlpr.ia.ac.cn}\\
{\tt\small zhuzheng2014@ia.ac.cn hongliang.bai@faceall.cn}
%\and
}
\renewcommand{\thefootnote}{\fnsymbol{footnote}} 

\maketitle

\begin{abstract}
%Visual object tracking is a challenging computer vision problem, which can fall in two groups: traditional correlation filters (CF) based method and end-to-end training convolutional neural network (CNN) framework.%
For visual tracking, most of the traditional correlation filters (CF) based methods suffer from the bottleneck of feature redundancy and lack of motion information. In this paper, we design a novel tracking framework, called multi-hierarchical independent correlation filters (MHIT). The framework consists of motion estimation module, hierarchical features selection, independent CF online learning, and adaptive multi-branch CF fusion. Specifically, the motion estimation module is introduced to capture motion information, which effectively alleviates the object partial occlusion in the temporal video. The multi-hierarchical deep features of CNN representing different semantic information can be fully excavated to track multi-scale objects. To better overcome the deep feature redundancy, each hierarchical features are independently fed into a single branch to implement the online learning of parameters. Finally, an adaptive weight scheme is integrated into the framework to fuse these independent multi-branch CFs for the better and more robust visual object tracking. Extensive experiments on OTB and VOT datasets show that the proposed MHIT tracker can significantly improve the tracking performance. Especially, it obtains a 20.1\% relative performance gain compared to the top trackers on the VOT2017 challenge, and also achieves new state-of-the-art performance on the VOT2018 challenge.

\end{abstract}

%%%%%%%%% BODY TEXT
\section{Introduction}

Visual object tracking is such a task that continually localizes and tracks a target in a video only given position information in the first frame. It has many real-world applications, such as automatic driving, robotic services, and object surveillance. However, it also faces some complex situations, such as foreground occlusions, illumination changes, and appearance changes. Therefore, how to design a robust tracker has drawn a significant amount of interest from both academia and industry.
\begin{figure}[t]
\begin{center}
\includegraphics[width= 8cm]{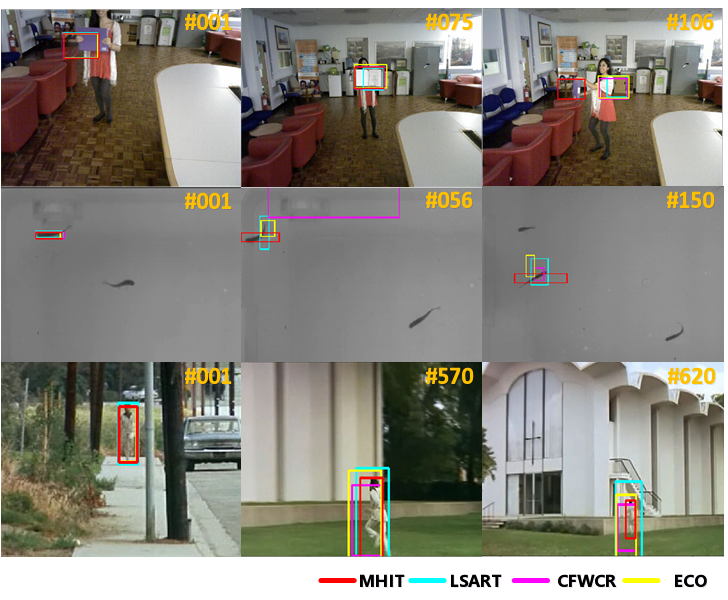}
   \caption{Example tracking results of different methods on the VOT2017 dataset. Our tracker MHIT(red) is more robust under deformation and scale changes. }
\end{center}
\end{figure}

Many attempts have been addressed to improve the performance of trackers in recent years. One group of methods mainly adopt deep features of CNN to train a tracker in end-to-end manner \cite{Siam,dcfnet,Li_2018_CVPR,zhu2018distractor}. This type of trackers obtains features with strong recognition ability by pretraining the CNN model offline on large-scale datasets (e.g., ImageNet  \cite{Russakovsky2015ImageNet} and Youtube-BB \cite{Real2017YouTube}). Another group is correlation filters (CF) based methods that mainly use the cyclic matrix to generate dense sampling for the online learning of CF parameters. The CF based trackers \cite{CCOT,ECO,Sun2018Learning,deepSRDCF} rely on strong feature representation with large number of parameters and frequent online updates. Researchers gradually shift attention from traditional handcrafted features (e.g., HOG \cite{dalal2005histograms}, Color Names (CN) \cite{van2009learning}) to more powerful multi-level CNN features.

\begin{figure*}[!ht]
\begin{center}
\includegraphics[width= 1\textwidth]{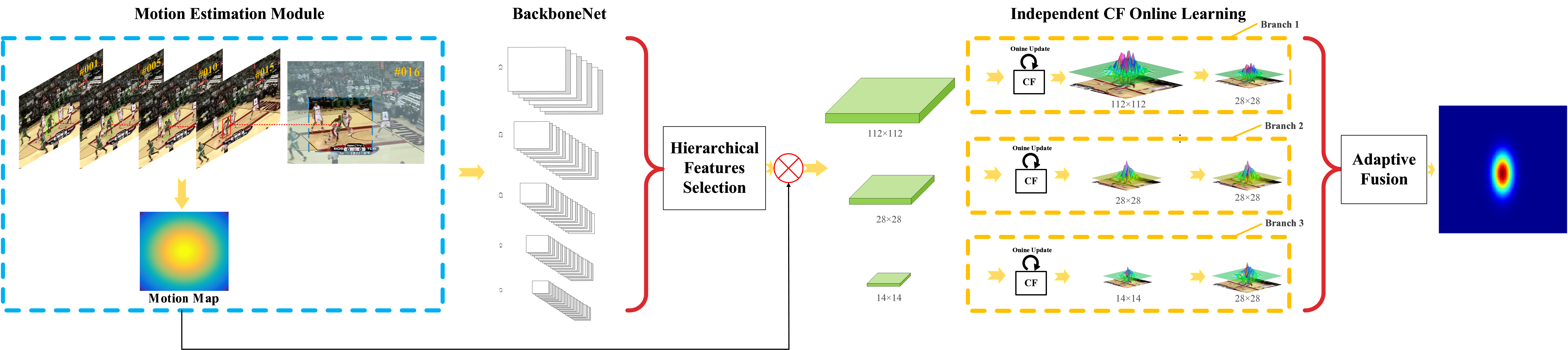}
   \caption{Overview of the architecture of our tracking framework, which consists of  motion estimation module, hierarchical features selection, independent CF online learning, and adaptive multi-branch CF fusion. The motion estimation module previously locates the target by learning the law of movement of the target according to historical frame and then determines the search area. Then hierarchical features are extracted by CNN to represent the hierarchical semantic information. After that, hierarchical  features are multiplied by the motion map generated by estimation module and then independently fed into different CFs to online update the parameters. Finally, an adaptive multi-branch CF fusion is utilized to generate the final score map to locate the center point coordinates of the target.}
   \label{fig2}
\label{fig:1}
\end{center}
\end{figure*}

However, most of the CF based trackers roughly concatenate all kinds of features, which can be defined as multiple features fusion, introducing much redundancy and burying the characteristics of hierarchical features. Moreover, a massive number of trainable parameters have improved the risk of severe over-fitting, which is indicated in ECO \cite{ECO}. Besides, due to the large parameters and sparse features of CNN, only limited hierarchical and shallow networks such as AlexNet \cite{Krizhevsky2012ImageNet} and VGG \cite{simonyan2014very} are used in the field of visual tracking over the past few years. Recently, the deeper network (e.g., ResNet \cite{He2016Deep}, DenseNet \cite{Huang2017Densely} and SE-ResNet \cite{Hu2017Squeeze} has shown the more efficient performance for many computer vision tasks. However, how to effectively utilize rich deep CNN features to construct a robust tracker is still a challenging study.

In this paper, we fully excavate multi-level deep convolutional features to form a more robust yet efficient tracking framework, called MHIT. Instead of the directly concatenating multi-level deep features in previous trackers, we independently use multi-hierarchical deep features to implement the online learning of different branch, and adaptively fuse the respective CF score map. To further improve the robustness of our framework, the motion estimation is introduced under the consideration of the time continuity for overcoming some difficult situations, such as complete occlusion and deformation.

In summary, we make the following contributions:
\begin{itemize}
\setlength{\itemsep}{0pt}
\setlength{\parsep}{0pt}
\setlength{\parskip}{0pt}
\item We propose a novel CF tracking framework, called MHIT, which efficiently fuses the multi-branch independent solutions of CF via an adaptive weight strategy for more robust and reliable tracking.
\item Different hierarchical features are independently fed into different branches to update parameters of CF online, which sufficiently alleviates the curse of dimensionality of the conventional multi-feature fusion. And a motion estimation is also addressed successfully to capture the motion information.
\item The proposed tracker achieves remarkable performance improvement on OTB2013 \cite{wu2013online}, OTB2015 \cite{wu2015object}, VOT2016 \cite{Matej2016The}, VOT2017 \cite{vot2017} and VOT2018 \cite{Kristan2018a} datasets. Especially, it achieves the new state-of-the-art on the VOT2017  and VOT2018  challenge.

\end{itemize}
%-------------------------------------------------------------------------
\section{ Related Work }

Based on the CNN methods, Li et al \cite{Li2015DeepTrack} exploit the CNN end-to-end training approach to turn the tracking problem into a classification problem. MDNet \cite{mdnet} further combines offline multi-domain training and online updates classifiers for identifying specific targets. Following the end-to-end ideas, some works further use a Siamese matching structure to learn a similarity measure, which regards DCF as part of the networks. SiamFC \cite{Siam} trains offline with the ILSVRC \cite{Russakovsky2015ImageNet} dataset and does not update the parameters online. DCFNet \cite{dcfnet} presents an end-to-end network architecture to learn the convolutional features and performs the correlation tracking process simultaneously. SiamRPN \cite{Li_2018_CVPR} introduces feature extraction and region proposal subnetwork including the classification branch and regression branch. DaSiamRPN \cite{zhu2018distractor} proposes a framework on the basis of SiamRPN \cite{Li_2018_CVPR} to learn distractor-aware features and explicitly suppress distractors during the inference of online tracking. SiamVGG \cite{Kristan2018a} replaces the base network AlexNet \cite{Krizhevsky2012ImageNet} with VGG \cite{simonyan2014very} on the basis of SiamFC \cite{Siam} to improve tracking performance. This type of methods typically takes the groundtruth of the first frame as a template or employs a simple moving average strategy to update the template.

As for the correlation filters algorithm, it has received extensive attention in visual tracking due to the high computational efficiency in the Fourier domain. Bolme et al. \cite{Bolme2010Visual} propose a CF tracker by learning a minimum output sum of squared error (MOSSE) for target appearance, which is able to run in high speed. CSK \cite{Henriques2012Exploiting} uses a circular matrix for dense sampling to generate a large number of samples with low computational load. KCF \cite{KCF} adopts ridge regression and multi-channel features to solve correlation filter parameters. SRDCF \cite{Danelljan2016Learning} makes use of a negative Gaussian penalty weight on the filter parameters to overcome the boundary effect. DeepSRDCF \cite{deepSRDCF} introduces CNN features into SRDCF \cite{Danelljan2016Learning} and achieves good results. C-COT \cite{CCOT} further converts feature maps of different resolutions into a continuous spatial domain to achieve better accuracy. The subsequent ECO \cite{ECO} improves the C-COT \cite{CCOT} tracker in terms of performance and efficiency. Based on ECO \cite{ECO}, CFWCR \cite{CFWCR} normalizes each individual feature extracted from different layers to get more robust results. Attempts on features design have shifted from the CN \cite{van2009learning}, HOG \cite{dalal2005histograms} hard-crafted features to CNN features.

\section{ Proposed Method }
\subsection{Correlation Filter for Visual Tracking}
We first review the traditional correlation filters algorithm. Each sample $x_{k}$ contains $D$ feature channels $x_{j}^1$ , $x_{j}^2$ ,..., $x_{j}^D$, extracted from the same image patch, where $k$ is the index of the samples. Assume that $f = \left \{ f_{d} \right \}_{d = 1: D}$ is a set of $D$ channel features. The correlation filters algorithm can be formulated as :
\begin{equation}
 \mathop{\arg\min}_{f}\sum_{k=1}^K||\phi(x_k,f)-y_k||_{L^2}^2+\lambda\sum_{d=1}^D||f||_{L^2}^2 ,
 \label{eq1}
\end{equation}
\begin{equation}
\phi(x_k,f) = \sum_{d=1}^Df_{d}*x^d_k,
\end{equation}
where $x_k$ is the cyclic shift sample of the $x_k$ and $y_{k}$ is the Gaussian response label. The optimization problem in Eq.\ref{eq1} can be solved efficiently in the Fourier domain.
 Eq.\ref{eq1} is minimized as $f=sum_{k=1}^K\alpha_kx_k$, where the coefficient $\alpha$ is computed with:
\begin{equation}
\alpha = \mathcal{F}^{-1}(\frac{\mathcal{F}(y)}{\mathcal{F}(\phi(x ,x)}+\lambda),
\end{equation}
where $\mathcal{F}$ and $\mathcal{F}^{-1}$ denote Fourier tranform and its inverse respectively.  Given $\alpha$ and the appearance model $\hat{x}$, we can get the response map  $\hat{y}$ of a new patch $z$ by:
\begin{equation}
\hat{y} = \mathcal{F}^{-1}(\mathcal{F}(\alpha)\odot \mathcal{F}(\phi(z,\hat x ))).
\end{equation}

\subsection{ Motion Estimation Module }
Most of existing DCF trackers only consider appearance features of current frame, and hardly benefit from motion information. The lack of temporal information degrades the tracking performance during challenges such as partial occlusion and deformation. Our proposed tracker uses motion estimation module to take full use of the motion information.

Kalman filtering is an algorithm that uses a series of measurements observed over time, which estimates a process by using a form of feedback control. The equations for Kalman filters fall in two groups: time update equations and measurement update equations. The time update equations can also be regarded as predictor equations, while the measurement update equations can be regarded as corrector equations. The time update projects the current state estimate ahead in time. The measurement update adjusts the projected estimate by an actual measurement at that time.

Time update equations can be formulated as:
\begin{equation}
\hat{x}_{k|k-1} = F_{k} \hat{x}_{k-1|k-1} + B_{k}u_{k},
\label{eq5}
\end{equation}

In Eq.\ref{eq5}, $\hat{x}_{k|k-1}$ is a vector representing predicted process state at time $k$ before measurement update, $u_{k}$ is a control vector and $B_{k}$ relates optional control vector $u_{k}$ into state space. $X$ is a 4-dimension vector $\begin{bmatrix}x& y& dx& dy\end{bmatrix}$ , where $x$ and $y$ represent the coordinates of the center of the target, and $dx$ and $dy$ represent its velocity. Therefore, process transition matrix $F_{k}$ can be expressed as:
\begin{equation}
F_{k} = \begin{bmatrix}
 1& 0&  1& 0\\
 0& 1&  0& 1\\
 0& 0&  1& 0\\
 0& 0&  0& 1
\end{bmatrix},
\end{equation}

The predictive estimated covariance matrix can be formulated as:
\begin{equation}
P_{k|k-1} = F_{k} P_{k-1|k-1}F_{k}^{T}+ Q_{k},
\end{equation}
where $P_{k|k-1}$ is the posterior estimate error covariance matrix which measures the accuracy of the estimate at time $k$ before measurement update and $Q_{k}$ is the process noise covariance at time $k$.

After time update process, the Kalman filter uses measurement to correct its prediction during the measurement update steps.
The measurement residual can be expressed as:
\begin{equation}
\hat{y_{k}} = z_{k} - H_{k}\hat{x}_{k|k-1},
\end{equation}
where $H_{k}$ is the matrix converting state space into measurement space at time $k$.
The measurement margin covariance can be expressed as:
\begin{equation}
S_{k} = H_{k}P_{k|k-1}H_{k}^{T}S_{k}+R_{k},
\end{equation}
where $R_{k}$ is measurement noise covariance at time $k$.
The optimal Kalman gain can be formulated as:
\begin{equation}
K_{k} = P_{k|k-1}H_{k}^{T}S_{k}^{-1},
\end{equation}

After that, we use the results of Eq.\ref{eq11} and Eq.\ref{eq12} to update the filter:
\begin{equation}
\hat{x}_{k|k} = \hat{x}_{k|k-1} + K_{k}\hat{y_{k}},
\label{eq11}
\end{equation}
\begin{equation}
\hat{P}_{k|k} = (I - K_{k}H_{k})P_{k|k-1},
\label{eq12}
\end{equation}

The motion of the object has a certain regularity, so the change in the size of the rectangle and the center coordinates can satisfy a certain law. The trajectory of the target generated by the DCF framework is not smooth enough to satisfy the motion law. After obtaining the final center point position by the DCF tracker, we use it as the observation of the current state to correction the Kalman filter. After that, we get a more accurate estimation with less noise.

\begin{figure}[h]
\begin{center}
\includegraphics[width=8cm]{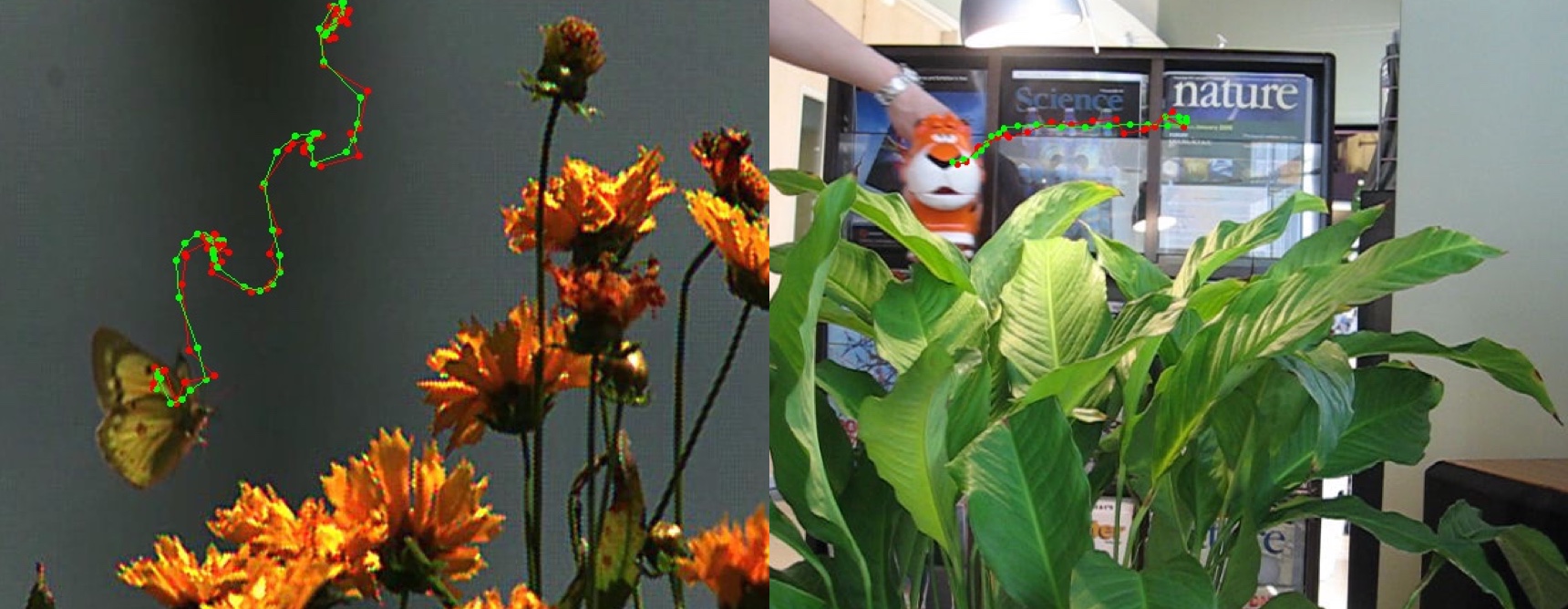}
   \caption{The figure shows the trajectory with motion estimation (green) and without motion estimation (red). As is shown by the figure, the trajectory becomes smoother with motion estimation due to less noise.}
\end{center}
\end{figure}

In the inference phase, we obtain the center point of the object by the motion estimation model and then we generate a cosine window as motion map centering on it. After that, we expand the patch to get the accurate search area to predict the location of the target. The Gaussian motion map is used to multiply the feature map generated by the CNN. Through the above steps, a more accurate estimation of the center point of the object is obtained, and the problem of center point drift caused by deformation and partial occlusion is alleviated.

\subsection{Hierarchical Features Selection}
Lin. et al \cite{Lin2016Feature} propose a novel feature pyramid networks (FPN), which uses top-down and lateral connections so as to capture the object with different scales. During online tracking, the size of the target changes frequently. Therefore, hierarchical features are introduced into our tracking framework. In this subsection, we investigate the specialty of different layers and the proper ways to combine them.

\begin{figure}[h]
\begin{center}
\includegraphics[width= 8cm]{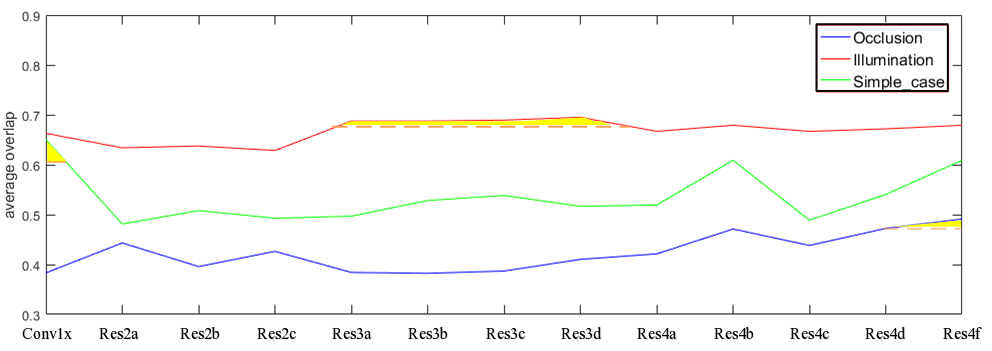}
   \caption{The performance of different layers in three tracking situations: occlusion, illumination and simple case. The $y$-axis utilizes the average overlap as the evaluation and the $x$-axis as the position of CNN.}
   \label{fig4}
\end{center}
\end{figure}

The semantic information of CNN layers from the shallow to the deep has specific performance to the tracking problem. The performance of different layers in three tracking situations (occlusion, illumination and simple case) is visualized in Figure \ref{fig4}. The shallower features provide rich details, which have better adaptability and precision. Yet, for objects with large deformation and occlusion, the performance goes worse. As for the middle layer features, there exist not only object outline but also powerful semantic information, which tends to be more stable for the scale variations. Deeper features have better stability when dealing with larger deformation. However, it is prone to drifting when having similar objects. Conventional multi-feature fusion directly combines the features of multiple levels to solve the problem, ignoring the interference between different levels, which causes failures under some complex situations.

Specifically, for ResNet \cite{He2016Deep}, we use the features before ReLU activations as output. The differences in features of adjacent layers are similar. Therefore, if we select adjacent layers, more redundancy and interference will be introduced into our tracking framework. We finally select the better performance levels of each stage in the graph as the feature extraction layers, including \textit{Conv1x}, \textit{Res3d} and \textit{Res4f}, which not only increases the specificity of features, but also can be easily adapted to dramatic scale variations.

\subsection{Independent Correlation Filters Online Learning}
Compared to the conventional CF algorithms, We treat the hierarchical features differently, and train a set of independent filters for each feature:

\begin{equation}x_l=\sum_i^{D_l}x_{l,i},\end{equation}
where $x_{l,i}$ is the feature of the $l_{th}$ layer and the $i_{th}$ channel. Each layer of a convolutional neural network can be viewed as a set of nonlinear filters. More complexity and redundancy will be introduced into the algorithm as the feature dimensions increase. Therefore, an input image is encoded by filters at each layer. If there are ${D}$ filters in one layer and the size of the feature map is $M \times M$, the number of channels corresponding to the feature map is also $D$ and the feature matrix $F$ belongs to $R^{D \times M \times M}$. With the deepening of the hierarchical features, $D$ increases gradually and the complexity of the obtained feature map also increases gradually. If we combine the multiple dimensional features directly, the dimension of the features $(\sum_i Di)$ will be high, which causes computational burden growing greatly. It is indicated to exist a lot of redundant information \cite{ECO}, which is not conducive to the solution of filters. Specific information can be independently learned by different training strategies. On the other hand, the adaptive multi-branch correlation filters fusion is able to improve the robustness effectively, which can be seen in the section 3.5.

Assume that $N$ layers of results are adopted, the correct probability of each result is $P_i$, and the probability of the correct result in the final results is between $Min\{P\}$ and $1-(1-p_i)^N$. When the correct rate of each result is above 0.7, the probability of having the correct result is between 0.7 and 1-$0.3^N$. And when the specificity among the $N$ results is more distinct, the eventual result tends to be 1-$0.3^N$. At the same time, as $N$ increases, $P$ tends to become 1. Therefore, it's efficient to improve the accuracy of the result by increasing the specificity in the selected levels or the number of layers. Instead, Too many layers will cause choosing the right result more difficult and encountering the overburden of many calculations. So the best solution is to compromise the specificity and the number of layers. We investigate CNN features applied to the exploration of differential performance in tracking problems in section 3.3.

The decomposed objective function can be expressed as $L$ independent solution objective functions:
\begin{equation}
\mathop{\arg\min}\sum_{k=1}^K||\phi(x_{l,k},f_l)-y_{l,k}||_{L^2}^2+\lambda\sum_{d=1}^{D_l}||w*f_{l,d}||_{L^2}^2,
\label{eq14}
\end{equation}
where $f_{l,d}$ is the $l_{th}$ layer and the $d_{th}$ channel filter parameters, and $y_k$ represents the predefined Gaussian window objective function.

For the above optimization problem in Eq.\ref{eq14}, we first solve the filter parameters $f$. As for each group of filters, we set the derivative to be zero, and the minimizer of Eq.\ref{eq14} is solved by the following normal equations, where $\Gamma = diag(f_{l,1},f_{l,2},...,f_{l,D_l}) \in R^{D_l\times D_l}$:
\begin{equation}A f= \Gamma ^TXy,\end{equation}
where $ A=\Gamma ^T X X^T\Gamma-\lambda W^T W$. Moreover, we adopt the Conjugate Gradient method to solve Eq.\ref{eq14} iteratively.

Most existing correlation filters algorithms tend to update at each frame, which causes high computational load. ECO \cite{ECO} updates the model in a fixed frame interval. It proves a sparser updating scheme is more efficient than the conventional strategy which updates every frame. By postponing update of the model a few frames, the loss is updated by adding a new mini-batch to the training samples, instead of only a single one, which helps to reduce over-fitting to the recent training samples. Intuitively, a sparser updating scheme leads to a low convergence speed. Hence, adopting more conjugate gradient iterations is necessary. More than that, to improve convergence rate, we choose a suitable momentum factor by using Fletcher-Reeves formula \cite{R1964Function} or the Polak-Ribiere formula \cite{Shewchuk1994An}.

\subsection{Adaptive Multi-branch Correlation Filters Fusion}

For multiple independent branches, the hierarchical filters are solved. Then we design an adaptive weight scheme to effectively fuse and obtain a more robust result. We call this weight ${\bf{m}},$ ${ \bf{m}}=[m_1,m_2,...,m_L]^T$. Then our final loss function can be formulated as follows:
\begin{equation}\mathop{E(f,m) = \sum_{l=1}^Lm_lE_l(f) + \sum_{l=1}^L||m_l||_{L^2}^2}_{s.t. \sum_{l=1}^Lm_l=1,m_l>=0},\end{equation}
we express the results of each layer as $E=[E_1,E_2,...,E_L]$. The optimization problem of $m$ can be converted to:
\begin{equation}
\mathop{\mathop{\arg\min}_{m} {\bf {\bf \large m}^TE+{\bf \large m}^T{\bf \large m}}}_{s.t. \sum_{l=1}^Lm_l=1,m_l>=0},
\label{eq17}
\end{equation}

Because of $\sum_{l=1}^Lm_l=1$, then $\sum_{l=1}^Lm_l\sum_d^{D_l} ||f_{l,d}||^2$ and $\sum_{l=1}^Lm_l||y_l||_{L^2}^2$ are constant, which can be ignored.
 $E_l$ can be converted to:
\begin{equation}E_l=C_l^TC_l-2C_l^Ty_l,\end{equation}
in this equation, $C_l=\mathcal{F}^{-1}(\sum_{d=1}^{D_l}X_{l,d}^H\Large \odot \mathcal{F}(f_{l,d}))$. Meanwhile, the problem in Eq.\ref{eq17} becomes a quadratic programming problem, which can be solved by standard quadratic programming.

The center point coordinates of the target can be obtained by the fusion score map, after which the scale of the target can be predicted. We apply a multi-scale search scheme, which takes the position predicted by the motion estimation module and takes multiple scales to extract the search area. In terms of our conclusion in section 3.3, medium-layer features are more robust for the determination of scales. We extract an image patch of size  $\alpha ^nP \times \alpha ^nR$ centered around the target, where $\alpha$ is 1.03, $n \in [-5,5]$. After that, we extract the medium features $(C_{scale})$ and employ filters to acquire response maps, where $scale$ is the position of CNN's layer, and $C_{scale}$ can be expressed as:
\begin{equation}
C_{scale} = \sum_{d=1}^{D_{scale}}\mathcal{F}^{-1}(\mathcal{F}(f_{scale,d})\odot (\mathcal{F}(x_{scale,d}))^H).
\end{equation}
\section{Experiment}
\begin{table*}
\begin{center}
\scalebox{0.9}{
\setlength{\tabcolsep}{1mm}{
\begin{tabular}{|c|c| c | c | c  |c|}
\hline
\multicolumn{2}{|c|}{ VOT2016} &\multicolumn{2}{|c|}{  VOT2017 } & \multicolumn{2}{|c|}{ VOT2018(H) }\\
\hline
Trackers & A \qquad  R \qquad  EAO &Trackers&  A \qquad R  \qquad EAO&Trackers&  A \qquad  R \qquad  EAO \\
\hline
\hline
DNT& 0.515\quad 0.329\quad 0.278 &MCPF&{ \color{blue}0.510}\quad 0.427\quad 0.248 &DLSTpp&  0.583\quad  0.454 \quad 0.196 \\
STAPLEp& 0.557\quad 0.329\quad 0.278 & SiamDCF& 0.500 \quad 0.473\quad  0.249&DASiamRPN&     {\color{red}0.628}\quad  0.518 \quad0.205 \\
SRBT &0.496\quad 0.350\quad 0.290 &CSRDCF& 0.491\quad  0.356\quad  0.256 & CPT&   0.577\quad  0.424 \quad0.209 \\
EBT   & 0.465\quad 0.252\quad 0.291 & CCOT& 0.494\quad  0.318\quad  0.267& DeepSTRCF&             0.600 \quad   0.444\quad 0.221 \\
DDC& 0.541\quad 0.345\quad 0.293 & MCCT&{\color{red} 0.525}\quad  0.323\quad  0.270& LADCF&    0.550\quad  0.375\quad0.222\\
Staple& 0.544\quad 0.378\quad 0.295 & Gnet& 0.502 \quad 0.276\quad  0.274 &RCO(Resnet) &   0.571\quad  {\color{blue}0.315}\quad   0.246\\
MLDF &0.490\quad 0.233\quad 0.311 & ECO& 0.483\quad  0.276\quad  0.280 &UPDT &{\color{blue}0.603}\quad 0.343\quad {\color{blue}0.247} \\
SSAT &{\color{blue}0.577}\quad 0.291\quad 0.321 & CFCF& 0.509 \quad 0.281\quad  0.286 &- &- \qquad - \qquad - \\
TCNN &0.554\quad 0.268\quad 0.325 &CFWCR& 0.484\quad  0.267\quad  0.303 &- &-  \qquad -  \qquad -\\
CCOT &0.539\quad {\color{blue}0.238}\quad {\color{blue}0.331} & LSART& 0.493\quad { \color{blue}0.218}\quad { \color{blue}0.323} &- &- \qquad -  \qquad - \\
\hline
\hline
{\bf {MHIT(Ours)}} &{\color{red}0.580}\quad {\color{red}0.111}\quad {\color{red}0.451} &{\bf {MHIT(Ours)}}&   {\color{blue}0.510} \quad {\color{red}0.138}\quad {\color{red} 0.388} &{\bf {MFT(Ours)}}&  0.577\quad  {\color{red}0.311}\quad  {\color{red}0.252} \\
\hline
\end{tabular}}}
\caption{Performance comparisons on VOT2016, VOT2017 and VOT2018 hidden sequences benchmarks. The red bold fonts and blue italic fonts indicate the best and the second best performance. When taking part in the VOT2018 challenge, we name our tracker MFT and RCO. }
\label{allresult}
%Our Se-ResNet50 version MFT and ResNet50 version RCO achieve the first place and third place respectively.
\end{center}
\end{table*}
\subsection{Implementation Details}
Our tracker is implemented on MATLAB using Matconvnet and AutoNN tools. We employ the \textit{Conv1x}, \textit{Res3d}, and \textit{Res4f} of ResNet50 and SE-ResNet50 as the layers of our feature extraction. To reduce and balance the dimensions of the hierarchical features, principal component analysis (PCA) is introduced into our tracking framework. Through PCA, the feature dimensions of \textit{Conv1x}, \textit{Res3d} and \textit{Res4f} are respectively reduced to 64, 256, 256. The search area range is set between 224*224 and 250*250. We utilize the same model update strategy as CFWCR \cite{CFWCR} and the maximum number of stored training samples is set to 50 to avoid over-fitting while the number of intermediate frames without training is set to 6. We select different Gaussian window variances for different layers, 1/12, 1/12, 1/3, from shallow to deep. All the parameters are determined by selecting a uniform validation set.

\subsection{Results on VOT}
The visual object tracking (VOT) challenge is a competition between short-term, model-free visual tracking algorithms which contains 60 sequences. For each sequence in the dataset, the tracker is evaluated by initializing by the rectangle of the target in the first frame. The toolkit will restart the tracker as long as the target is lost. The robustness is obtained by counting the average number of failures, and the accuracy is the statistical average crossover ratio.

Table \ref{allresult} shows accuracy (A) and robustness (R), as well as expected average overlap (EAO) on VOT2016 \cite{Matej2016The}, VOT2017 \cite{vot2017} and VOT2018 \cite{Kristan2018a} hidden datasets.

\subsubsection{Results of VOT2016}
 The results in Table \ref{allresult} are presented in terms of expected average overlap (EAO), robustness (R), and accuracy (A). For clarity, we show the comparison with the top-10 best trackers, including CCOT \cite{CCOT}, TCNN \cite{tcnn}, SSAT \cite{Matej2016The}, MLDF \cite{Matej2016The}, Staple \cite{staple}, DDC \cite{Matej2016The}, EBT \cite{Zhu2016Beyond}, SRBT\cite{Matej2016The}, STAPLE+ \cite{staple} in the VOT2016 \cite{Matej2016The} competition.
\begin{figure}[h]
\begin{center}
\includegraphics[width=8cm]{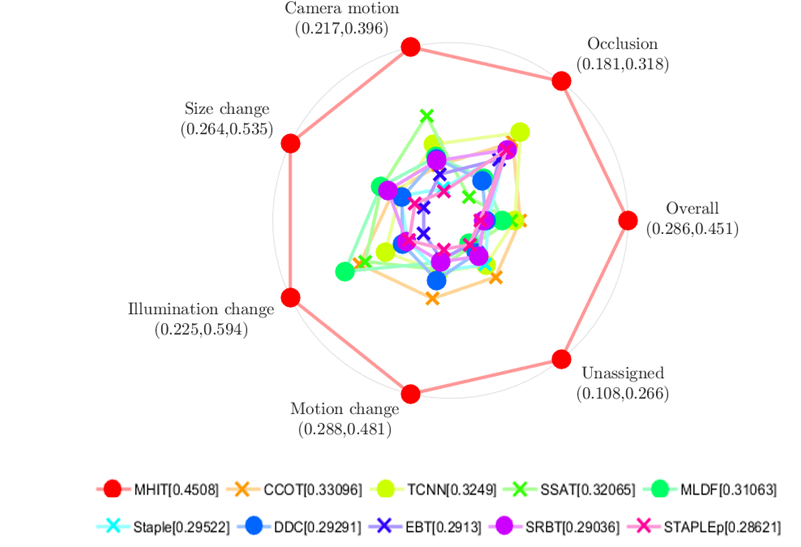}
   \caption{Per-attribute plot for ten top-performing trackers on VOT2016 in EAO. Best viewed on color display.}
   \label{figvot16}
\end{center}

\end{figure}

 Our proposed MHIT outperforms all top 10 trackers at the EAO score 0.451. The MHIT significantly  precedes CF approaches that do not apply deep CNN and end-to-end training CNN based trackers. Meanwhile, Figure \ref{figvot16} shows per-attribute plot for ten top-performing trackers on VOT2016 in EAO. In all the attributes (size change, camera motion, occlusion, unassigned, motion change, illumination change), our MHIT tracker gets better performance than other state-of-the-art results, which also demonstrates the effectiveness of our tracking framework.

\subsubsection{Results of VOT2017}
We compare the tracking results \cite{vot2017} of the top 10 trackers, including LSART \cite{Sun2018Learning}, CFWCR \cite{CFWCR}, CFCF \cite{gundogdu2018good}, ECO \cite{ECO}, Gnet \cite{vot2017}, MCCT \cite{vot2017}, CCOT \cite{CCOT}, CSRDCF \cite{CSRDCF}, SiamDCF \cite{vot2017} on the VOT2017 challenge. LSART \cite{Sun2018Learning} achieved the first in the VOT2017 public 60 sequences with the EAO of 0.323. By fully excavating different sematic information, the proposed MHIT framework achieves a relative gain of 20.1\% compared to LSART in EAO.

\begin{figure}[h]
\begin{center}
\includegraphics[width= 9cm]{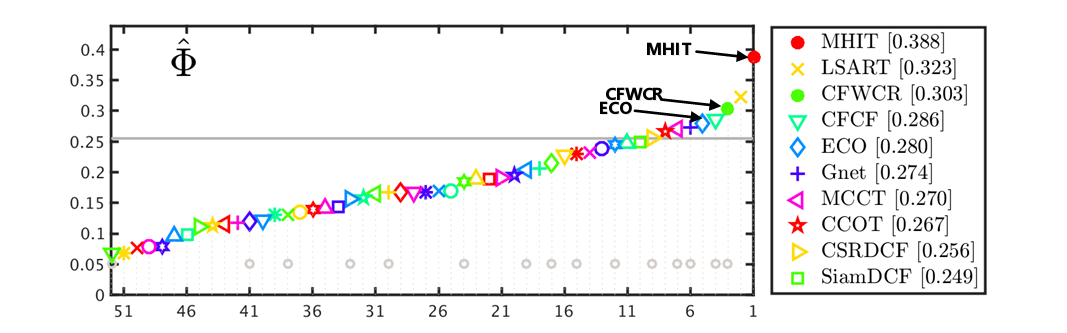}
   \caption{EAO on VOT2017. This figure only shows the plots of top 10 trackers and our MHIT for clarity. }
\end{center}
\end{figure}

\subsubsection{Results of VOT2018 }
The public 60 sequences of VOT2018 \cite{Kristan2018a} public remain unchanged compared to VOT2017. Besides, VOT2018 also adopts EAO, accuracy (A) and robustness (R) for evaluations. The official top 8 trackers, include MHIT, UPDT \cite{bhat2018unveiling}, RCO \cite{Kristan2018a}, LADCF \cite{Xu2018Learning}, DeepSTRCF \cite{Kristan2018a}, CPT \cite{Kristan2018a}, DaSiamRPN \cite{zhu2018distractor,Li_2018_CVPR}, DLSTpp \cite{Kristan2018a} in the hidden dataset.

Table \ref{allresult} illustrates that our MHIT ranks 1st according to EAO criterion. When taking part in the VOT2018 challenge, we name our tracker MFT(SE-ResnNet50) and RCO(ResNet50). The robustness of our tracker is 9.4\% better than that of the other trackers. The robustness and the EAO of our tracker outperform all the state-of-the-art trackers in the VOT2018 challenge.

\subsection{Results on OTB}

\subsubsection{Results of OTB2013}
The OTB2013 dataset \cite{wu2013online} is one of the most widely used canonical dataset in visual tracking, which contains 50 image sequences with various challenging factors. The evaluation is based on two metrics: precision plot and success plot. Mean overlap precision (OP) is defined as the percentage of frames in a video where the intersection-over-union overlap exceeds a threshold of 0.5. The area under curve (AUC) of each success plot is used to rank the tracking algorithm. AUC is computed from the success plot, where the mean OP over all videos is plotted over the range of thresholds [0, 1]. To reduce clutter in the graphs, we show only the results for top-performing recent baselines, i.e., ECO \cite{ECO}, CCOT \cite{CCOT}, SRDCFdecon \cite{Danelljan2016Adaptive}, DeepSRDCF \cite{deepSRDCF}, SRDCF \cite{Danelljan2016Learning}, siamfc3s \cite{Siam}, Staple \cite{staple}.
\begin{figure}[h]
\begin{center}
\includegraphics[width= 8cm]{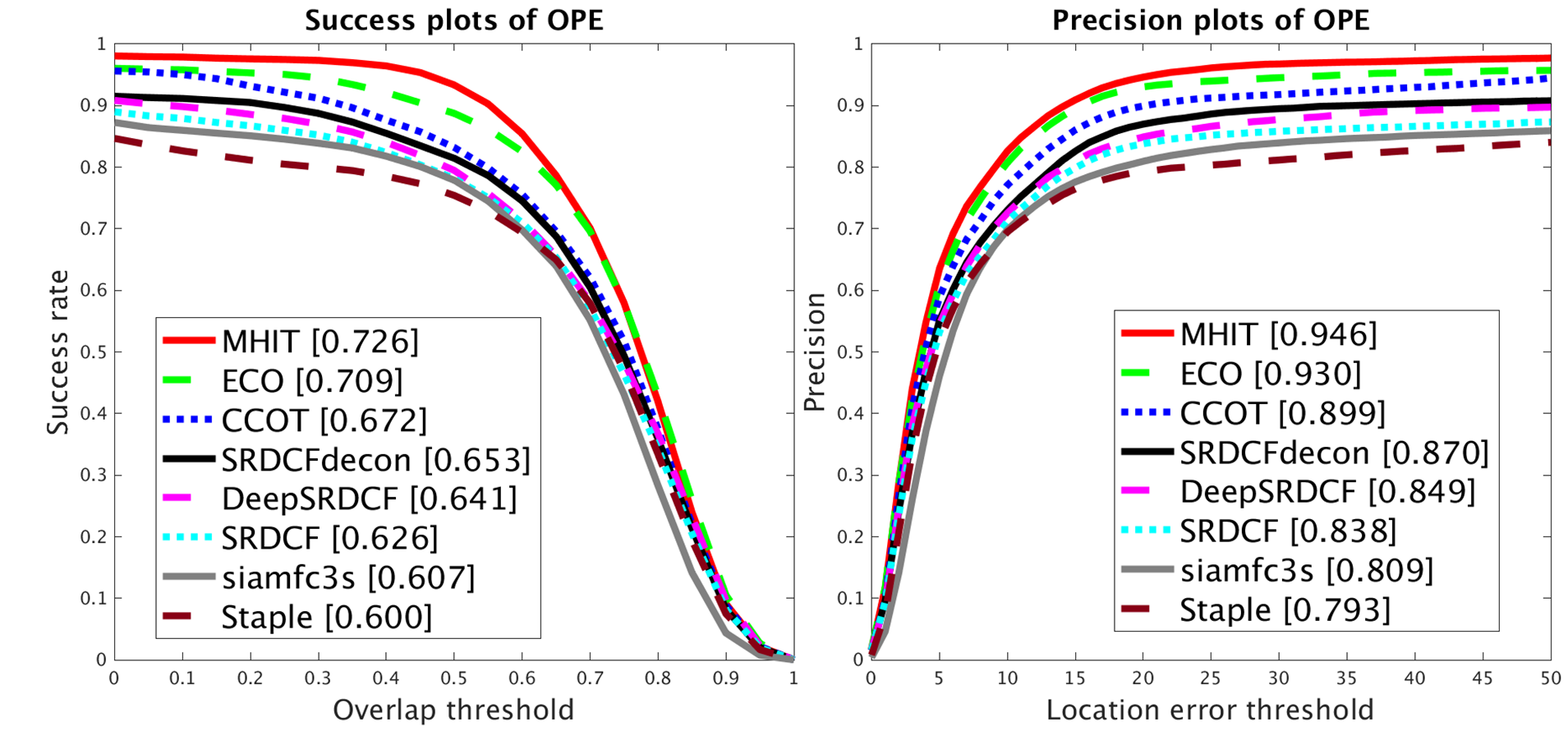}
   \caption{OTB2013 benchmark comparison. The precision plot (left) and the success plot (right).}
\end{center}
\end{figure}

Among all compared trackers, the proposed MHIT method obtains the best performance, which achieves the 94.6\% distance precision rate at the threshold of 20 pixels and a 72.6\% area-under-curve (AUC) score. Performance evaluation on different attributes of OTB2013 can be found in supplement material.

\subsubsection{Results of OTB2015}
The OTB2015 \cite{wu2015object} dataset is an extension of the OTB2013 dataset, which contains 50 more video sequences. We also evaluate the performance of the proposed MHIT method over all 100 videos in this dataset. To reduce clutter in the graphs, we show only the results for top-performing recent baselines, i.e., ECO \cite{ECO}, CCOT \cite{CCOT}, SRDCFdecon \cite{Danelljan2016Adaptive}, DeepSRDCF \cite{deepSRDCF}, SRDCF \cite{Danelljan2016Learning}, siamfc3s \cite{Siam}, Staple \cite{staple}.
\begin{figure}[h]
\begin{center}
\includegraphics[width = 8cm]{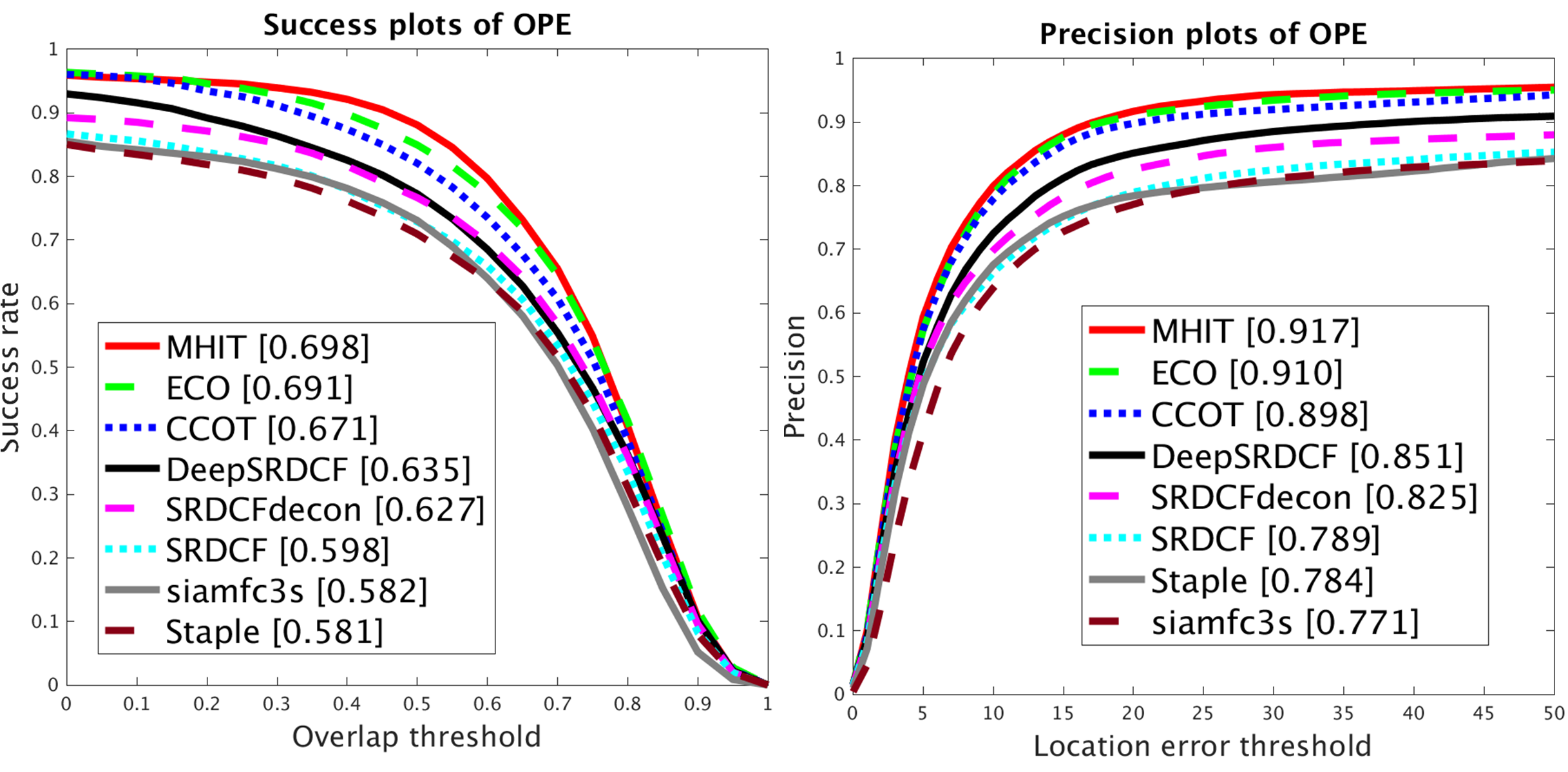}
   \caption{OTB2015 benchmark comparison. The precision plot (left) and the success plot (right).}
\end{center}
\end{figure}

Overall, our MHIT method provides the best result with a distance precision rate of 91.7\% and with an AUC score of 69.8\%, which again achieves a substantial improvement of several out-standing trackers (e.g., ECO, C-COT and DeepSRDCF). Performance evaluation on different attributes of OTB2015 can be found in supplement material.

\subsection{Ablation Analyses}

In this subsection, ablation analyses are performed to illustrate the effectiveness of proposed components. To verify the contributions of each component in our algorithm, the variations of our approach are implemented and evaluated.

\begin{figure}[h]
\begin{center}
\includegraphics[width=7cm]{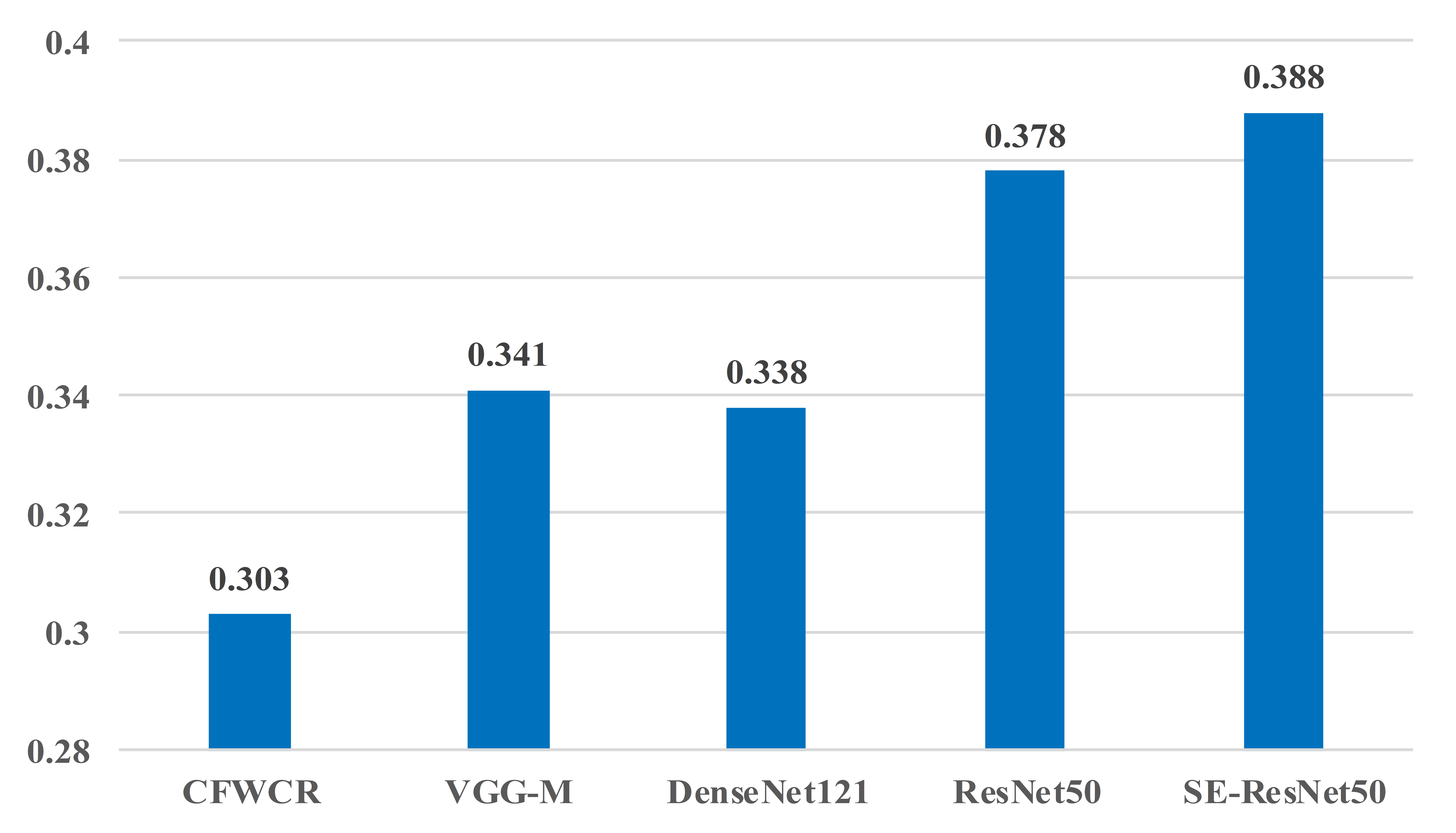}
   \caption{Tracking performance on VOT2017 when using deep features extracted from different networks.}
   \label{fig9}
\end{center}
\end{figure}

\textbf{Feature Comparisons} We compare the performance of VGG-M \cite{simonyan2014very}, Densenet121 \cite{Huang2017Densely}, ResNet50 \cite{He2016Deep} and SE-ResNet50 \cite{Hu2017Squeeze} in Figure \ref{fig9}. In all cases, we employ the same shallow representation, consisting of HOG and Color Names. The baseline CFWCR does not benefit from deeper and more powerful ResNet backbone. When using our MHIT framework, the EAO comes to 0.341 with the original VGG-M backbone, which demonstrates the effectiveness of independent correction filters. When using the deeper Resnet50 and SE-Resnet50 backbone in our proposed tracking framework, the power of CNN is significantly unveiled. In conclusion, our approach is able to exploit more powerful representations, which achieves a remarkable gain going from hand-crafted features towards more powerful network architectures.

After that, we compare the performance of different combinations of features extracted by SE-ResNet50, which is shown in Table \ref{tab_com}. If we select adjacent layers, more redundancy and interference will be introduced into our tracking framework, thus causing the performance degradation. Besides, according to the Table \ref{tab_com}, \textit{Conv1x} with the spatial size of 112$\times$112 provides richer details of target outline than \textit{Res2a} with the spatial size of 56$\times$56. The middle layer of SE-ResNet50, \textit{Res3d} can provide not only object outline but also powerful semantic information, thus significantly improving the performance.

\begin{table}
\begin{center}
\scalebox{0.7}{
\setlength{\tabcolsep}{2mm}{
\begin{tabular}{c c c c c c|c}
\hline
{\normalsize{Conv1x}} & {\normalsize{Res2a}} & {\normalsize{Res2c}} & {\normalsize{Res3d}} & {\normalsize{Res4a}} & {\normalsize{Res4f}} & \\
{\normalsize{(112$\times$112)}}&{\normalsize{(56$\times$56)}}&{\normalsize{(56$\times$56)}}&{\normalsize{(28$\times$28)}}&{\normalsize{(14$\times$14)}}&{\normalsize{(14$\times$14)}}& {\large{EAO}}\\
\hline
{\LARGE{$\checkmark$}}& & & & & & {\large{0.192}}\\
 & & &{\LARGE{$\checkmark$}} & & & {\large{0.253}}\\
 & & & & & {\LARGE{$\checkmark$}}& {\large{0.254}}\\
{\LARGE{$\checkmark$}}& & & & &{\LARGE{$\checkmark$}}  & {\large{0.284}}\\
& {\LARGE{$\checkmark$}}& & {\LARGE{$\checkmark$}}& &{\LARGE{$\checkmark$}} & {\large{0.357}}\\
{\LARGE{$\checkmark$}}& & & {\LARGE{$\checkmark$}}& &{\LARGE{$\checkmark$}}  & {\large{\color{red}0.388}}\\
& {\LARGE{$\checkmark$}}& & {\LARGE{$\checkmark$}}&{\LARGE{$\checkmark$}} &{\LARGE{$\checkmark$}}  & {\large{0.325}}\\
{\LARGE{$\checkmark$}}& & {\LARGE{$\checkmark$}}& {\LARGE{$\checkmark$}}& &{\LARGE{$\checkmark$}}  & {\large{0.339}}\\
{\LARGE{$\checkmark$}}& & & {\LARGE{$\checkmark$}}&{\LARGE{$\checkmark$}} &{\LARGE{$\checkmark$}}  & {\large{0.308}}\\
\hline
\end{tabular}}}
\end{center}
 \caption{Tracking performance of different combinations of SE-ResNet50 features.}
 \label{tab_com}
\end{table}

\begin{figure}[h]
\begin{center}
\includegraphics[width=8.5cm]{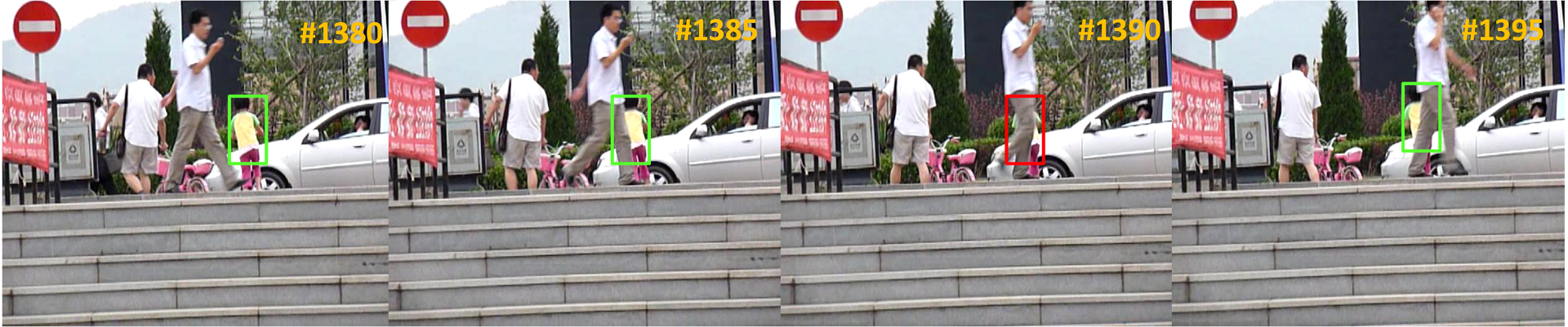}
   \caption{ Qualitative example of our proposed method with motion estimation module.}
   \label{diffnet}
\end{center}
\end{figure}
\textbf{Module Analyses} We next investigate the effect of proposed four modules, which can be visualized Table \ref{comself}. Baseline means using VGG-M without independent correlation filters, IS+VGG-M means introducing independent solution into the baseline, and fusing by adding with same weights. IS+ResNet50 means replacing VGG-M with ResNet50 as backbonenet. IS+AMF+ResNet50 joins the adaptive multi-branch fusion method. To verify the importance of our feature diversity and feature quantity in Section 3.4, we further add the feature extracted from the SE-ResNet50 on the basis of IS+AMF+ResNet50, called IS+AMF+SE-ResNet50. Besides, IS+AMF+SE-ResNet50+ME means adding motion estimation module on the basis of IS+AMF+SE-ResNet50. According to Table \ref{comself}, each module is compared with the baseline, improving respectively 3.4\%, 6.2\%, 7.3\%, 8.3\%, 8.5\% in terms of EAO.

The tracking performance under partial occlusion is visualized in Figure \ref{diffnet}. When the girl in the figure suffers from short-term partial occlusion (marked in red), motion estimation can help the tracking framework to predict the position of the target.

% \begin{figure}[t]
% \begin{center}
% \includegraphics[height=4.5cm]{compare_self.png}
% \caption{Expected Average Overlap (EAO) curve for different strategies on the VOT2017 dataset.}
% \label{com_sel_fig}
% \end{center}
% \end{figure}

\begin{table}
\begin{center}
\scalebox{0.9}{
\begin{tabular}{c c c c }
\hline
&EAO & A&R\\

\hline
Baseline& 0.303&0.484&0.267\\
 IS+VGG-M &0.337&0.493&0.205\\
 IS+ResNet50  &0.365&0.495&0.175\\
 IS+AMF+ResNet50&0.375&0.507&0.155\\
 IS+AMF+SE-ResNet50 &0.385&0.505&0.140\\
 IS+AMF+SE-ResNet50+ME &\color{red}0.388&\color{red}0.510&\color{red}0.138\\

%  &Baseline & IS+VGG\_M & IS^{Res50} & IS+AMF^{Res50} & IS+AMF^{SE_Res50} &
% IS+AMF+ME^{SE_Res50}\\
% EAO &0.303&0.337&365&0.375&0.385&\color{red}0.388\\
% A &0.484&0.493&0.495&0.507&0.505&\color{red}0.510\\
%  R&0.267&0.205&0.175 &0.155&0.140&\color{red}0.138\\
\hline

\end{tabular}}

\end{center}
 \caption{The table shows expected average overlap (EAO), as well as accuracy and robustness raw values (A,R) for different strategies.}
 \label{comself}
\end{table}

\section{Conclusion}
In this paper, we fully utilize hierarchical features and multi-branch correlation filters fusion to construct a novel CF based tracking framework, which efficiently remits curse of dimensionality of conventional multi-feature fusion. Motion estimation module is introduced into the framework as a supplement to the appearance information so as to capture the motion state of the target and remit the partial occlusion. Besides, our tracking framework benefits from the online learning to adapt appearance changes and scale variances in continuous image sequences. Finally, extensive experiments verify the efficiency of the proposed tracker, which achieves new state-of-the-art performance on both OTB and VOT benchmarks.

{\small
\bibliographystyle{ieee}
\bibliography{egbib}
}

\clearpage
\section{Supplementary Material}

\subsection{Detailed results on OTB2013}

\begin{figure*}[ht]
\includegraphics[width= 0.9\textwidth]{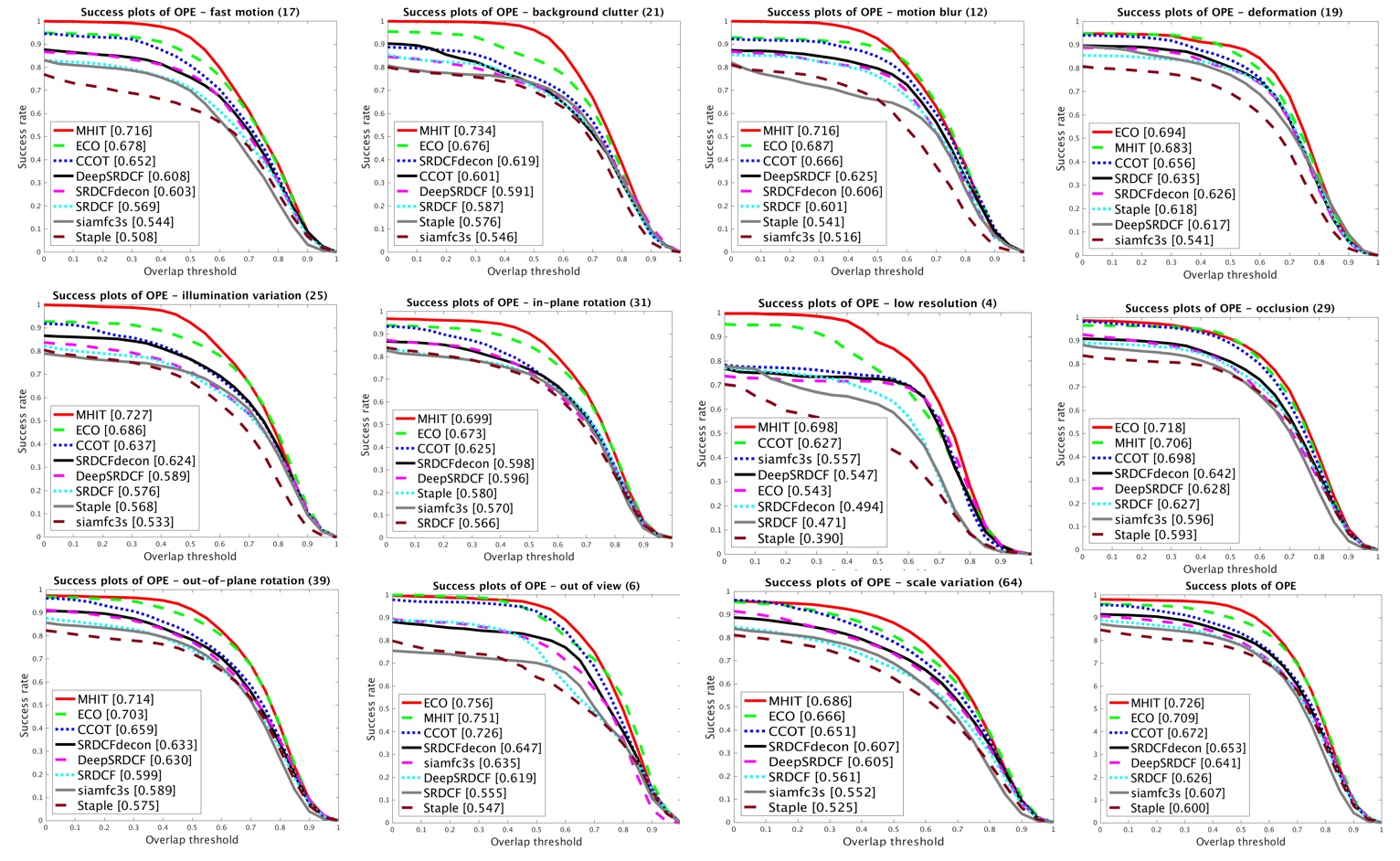}
   \caption{The success plots on OTB2013 for eleven challenge attributes.}
\label{fig_otb2013}
\end{figure*}

In this subsection, detailed results on OTB2013 are provided. Figure \ref{fig_otb2013} shows the success plots for all 11 attributes, including abrupt motion, background clutter, blur, deformation, in-plane rotation, low resolution, illumination variation, occlusion, out-of-plane rotation, out-of-view and scale variation on OTB2013.

Our tracker MHIT obtains remarkable performance with good robustness, which outperforms state-of-the-art tracker ECO\cite{ECO} in most of the attributes. In the evaluation of attributes of fast motion and motion blur, MHIT achieves 5.6\% and 4.2\% relative AUC gain compared to ECO\cite{ECO}, respectively. It illustrates the effectiveness of motion estimation module, which captures motion information to pre-locate the position of the target and generates a motion map to rectify the final score map. Moreover, due to the motion estimation module, our tracker also outperforms ECO\cite{ECO} under deformation and occlusion issues. Benefits from the powerful multi-hierarchical deep features of SE-ResNet50 and independent correlation filters, in the cases of scale variation, MHIT achieves a 3.0\% relative AUC gain compared to ECO\cite{ECO}.

\subsection{Detailed results on OTB2015}
\begin{figure*}[h]
\includegraphics[width= 0.9\textwidth]{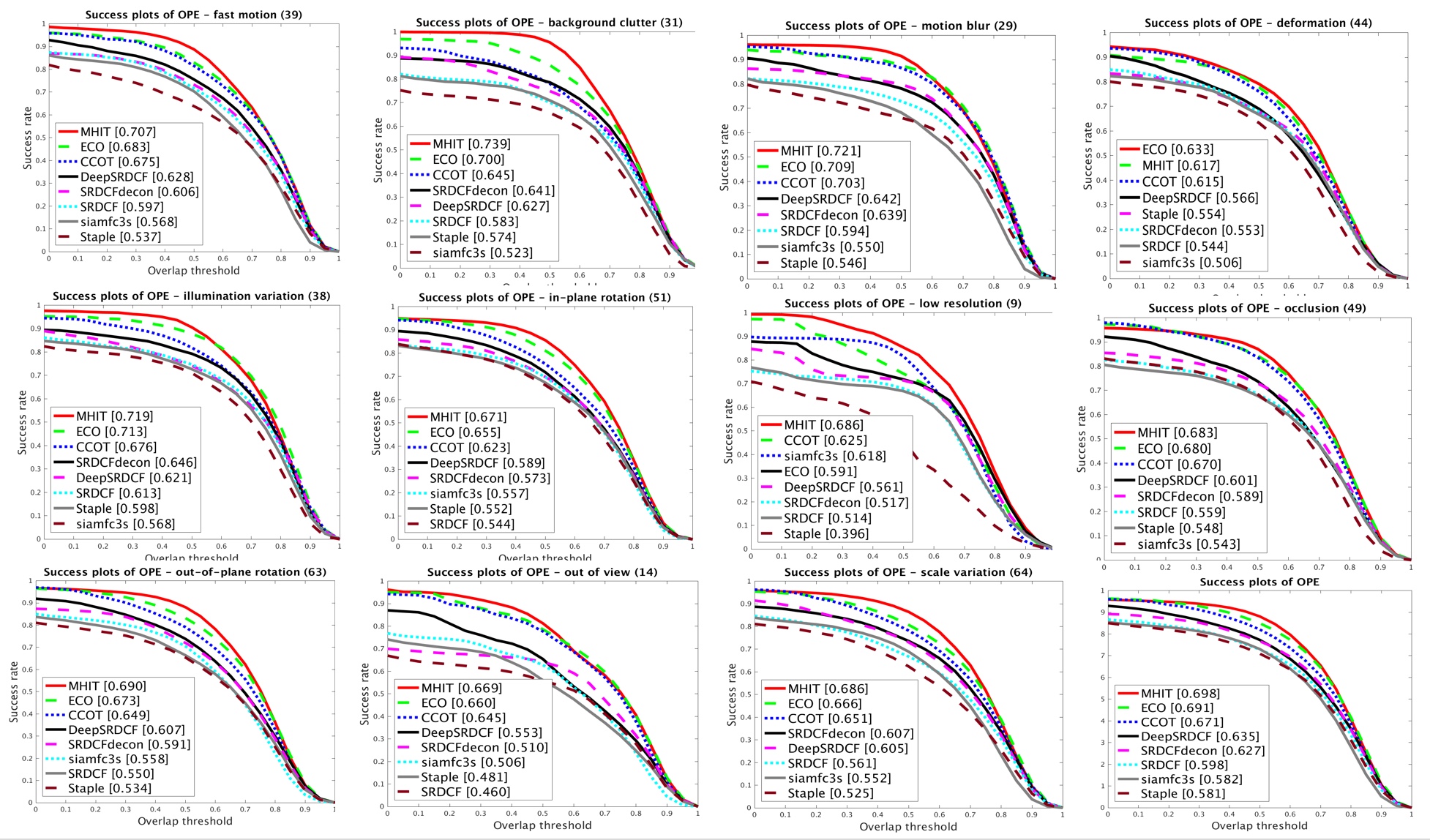}
   \caption{The success plots on OTB2015 for eleven challenge attributes.}
\label{fig_otb2015}
\end{figure*}

In this subsection, we provide detailed results on OTB2015. Figure \ref{fig_otb2015} shows the success plots for all 11 attributes, including abrupt motion, background clutter, blur, deformation, in-plane rotation, low resolution, illumination variation, occlusion, out-of-plane rotation, out-of-view and scale variation on OTB2015. Our tracker MHIT also significantly outperforms ECO\cite{ECO} in most of the attributes, which shares consistent results with OTB2013.

\clearpage

\end{document}